%% file: 0.MAIN.tex
\documentclass{IEEE_lsens}

\usepackage{textcomp}
\usepackage{soul}
\usepackage{amsthm,amsfonts,amssymb}
\usepackage{algorithm} 
\usepackage{algpseudocodex}
\usepackage{graphicx}
\usepackage{textcomp}
\usepackage{optidef}

\usepackage{soul}
\usepackage{url}

\usepackage{enumitem}
\usepackage{siunitx}
\usepackage{subfig}

\usepackage{multirow}
\usepackage{tikz}
\usepackage{import}
\usetikzlibrary{calc}
\usetikzlibrary{pgfplots.groupplots}
\usepackage{pgfplots}
\usepackage{pgfplotstable}
\usepackage{graphics}
\usepgfplotslibrary{fillbetween}
\usetikzlibrary{patterns}
\usepackage[export]{adjustbox}
\usepackage{pgfplotstable}

\definecolor{darkgray176}{RGB}{176,176,176}
\definecolor{darkturquoise0191191}{RGB}{0,191,191}
\definecolor{green01270}{RGB}{0,127,0}
\definecolor{lavender204204255}{RGB}{204,204,255}
\definecolor{lemonchiffon255255204}{RGB}{255,255,204}
\definecolor{lightgoldenrodyellow229255204}{RGB}{229,255,204}
\definecolor{lightgray204}{RGB}{204,204,204}

\usetikzlibrary{shapes.geometric}
\pgfdeclareplotmark{mystarc}{
    \node[star,star point ratio=2.25,minimum size=6pt,
          inner sep=0pt,fill= lemonchiffon255255204, draw = orange!55!white] {};
}

\usetikzlibrary{shapes.geometric}
\pgfdeclareplotmark{mystara}{
    \node[star,star point ratio=2.25,minimum size=6pt,
          inner sep=0pt,fill= darkturquoise0191191, draw = blue!55!white] {};
}

\usepackage[noadjust]{cite}

\ifCLASSINFOpdf
\else
\fi

\usepackage[T1]{fontenc} 
\usepackage{amsmath}
\usepackage[cmintegrals]{newtxmath}

\usepackage{bm}

\usepackage{array}

\usepackage{url}

\ifCLASSINFOpdf
\else
\fi

\providecommand{\hypersetup}[1]{\relax}

\hypersetup{pdftitle={Bare Demo of IEEE\_lsens.cls for IEEE Sensors Letters},
pdfsubject={Typesetting},
pdfauthor={Michael D. Shell},
pdfkeywords={Class, IEEE, IEEE\_lsens, IEEE Sensors Letters, LaTeX, Typesetting, TeX}}

\usepackage{tikz}

\begin{document}

\markboth{Vol.~1, No.~3, July~2017}{0000000}

\IEEELSENSarticlesubject{TinyML, Hardware Aware Neural Architecture Search, Ultra-Low-Power Computing}

\title{An affordable hardware-aware neural architecture search for deploying convolutional neural networks on ultra-low-power computing platforms}

\author{\IEEEauthorblockN{Andrea~Mattia~Garavagno\IEEEauthorrefmark{1}\IEEEauthorrefmark{2}, Edoardo~Ragusa\IEEEauthorrefmark{1}\IEEEauthorieeemembermark{1}, Antonio~Frisoli\IEEEauthorrefmark{2},
and~Paolo~Gastaldo\IEEEauthorrefmark{1}}
\IEEEauthorblockA{\IEEEauthorrefmark{1}Dept. of Electrical, Electronic, Telecommunications Engineering and Naval Architecture (DITEN),
University of Genoa, Genoa, Italy\\
\IEEEauthorrefmark{2} 
Department of Excellence in Robotics \& AI and Institute of Mechanical Intelligence, Scuola Superiore Sant’Anna, 56127 Pisa, Italy \\
\IEEEauthorieeemembermark{1}Member, IEEE}%
\thanks{Corresponding author: Andrea~Mattia~Garavagno (AndreaMattia.Garavagno@\{edu.unige.it, santannapisa.it).\protect\\}

}

\IEEELSENSmanuscriptreceived{Manuscript received ;
revised ; accepted .
Date of publication ; date of current version .}

\IEEEtitleabstractindextext{%
\begin{abstract}
Hardware-aware neural architecture search (HW-NAS) allows the integration of Convolutional Neural Networks (CNNs) in microcontrollers devices by automatically designing neural architectures that can fit prearranged hardware constraints. However, state-of-the-art HW-NAS target high-performance microcontrollers, whose power consumption does not meet sensing nodes requirements. This work presents a HW-NAS generating tiny CNNs that can run on ultra-low-power microcontrollers, featuring a lightweight search procedure enabling its execution even on embedded devices. Empirical results on three well-known benchmarks for tiny computer vision proved that the proposed HW-NAS was able to generate tiny CNNs while preserving state-of-the-art classification accuracy. 
\end{abstract}

}


\maketitle

\begin{tikzpicture}[remember picture,overlay]
\node[anchor=south,yshift=15pt] at (current page.south) {
    \parbox{\textwidth}{
        \centering \scriptsize
        This work has been accepted for publication in IEEE Sensors Letters. Final publication is available at \url{https://doi.org/10.1109/LSENS.2024.3387056}.
    }
};
\end{tikzpicture}

\begingroup\renewcommand\thefootnote{\textsection}
\endgroup
\input{1.intro}

\input{2.related}

\input{3.proposal}

\input{4.exp}

\section{Conclusion}
Many applications can profit from the integration of deep learning into sensing nodes; 
however, design and deployment require domain-specific skills and specialized computing resources. We proposed a Hardware aWare Neural Architecture Search (HW-NAS) for ultra-low-power MCUs supported by a lightweight optimization procedure. This NAS proved able to fit the constraints of different target devices while providing state-of-the-art CNNs on the Visual Wake Words dataset. In addition, the search procedure can run in a few hours without requiring GPUs. Experiments indeed showed that the HW-NAS can even run on a Raspberry Pi 4. 

A few aspects need further investigation. First, large datasets may represent an issue, in particular, if the search procedure should run on an embedded system. Second, as the proposed HW-NAS is designed for tiny CNNs, it may not achieve state-of-the-art results when the target platforms belong to high-performance MCUs. In this case, traditional NAS or hand-crafted architectures may provide valuable solutions in terms of generalization performance.

\vspace{-6pt}


\section*{Acknowledgment}
\addcontentsline{toc}{section}{Acknowledgment}
\scriptsize
 Project funded under the National Recovery and Resilience Plan (NRRP), Mission 4 Component 2 Investment 1.1 - Call for tender No. 1049 published on Sept 14, 2022 by the Italian Ministry of University and Research (MUR) funded by the European Union – NextGenerationEU. Project Title "LEARN - muLtimodal Edge computing-bAsed weaRable exoskeletoNs for assistance in daily life" – CUP D53D23016200001- Grant Assignment Decree No. 1181 adopted on July 27, 2023 by MUR.

\normalsize
\vspace{-6pt}

%
%

\bibliographystyle{IEEEtran}
\bibliography{refs2}


\end{document}

%% file: 1.intro.tex
\section{Introduction}
\label{sec:introduction}
\vspace{-5pt}
Sensing nodes can benefit from real-time inference procedures supported by machine learning algorithms. For example, teleceptive sensing, i.e., sensing without physical contact (e.g., optical sensors, radars, or depth sensors), strongly relies on convolutional neural networks (CNNs) \cite{ragusa2023affordance}.
CNNs, though, bring about high computing costs and non-trivial training procedures. These features are not compliant with sensing nodes that must deal with severe energy constraints \cite{gidon2023bi}. Hardware-aware Neural Architecture Search (HW-NAS) is an appealing solution because it provides an automatic procedure that generates and trains suitable CNNs given the constraints that characterize the physical device running the inference phase.

The existing HW-NAS \cite{Micronets,MCUNet,liberis2021munas} generates tiny CNNs suitable for execution in high-end microcontroller units (MCUs) empowered with large memories; indeed, they in general involve resource-demanding optimization procedures. Two distinct qualities would allow a broader utilization of such techniques in sensing nodes: 1) the ability of generating tiny CNNs compliant with tighter hardware constraints (e.g. when the target is a ultra-low power MCU) and 2) a limited computation cost of the NAS procedure. In practice, a procedure requiring hundreds of hours of GPU computing may not be suitable in some applications. For example,  a gateway producing custom CNNs for IoT edge nodes on local data, without accessing to the cloud, needs a HW-NAS so light that can run on embedded devices.

Battery-operated nodes often rely on ultra-low-power MCUs as computing engine; near-sensor computing \cite{borah2023aicarebreath} is a very good example. Recent works \cite{garavagno2024colabnas,garavagno2023hardware,garavagno2024running} proved that NAS techniques can be exploited for the design of CNNs that comply with the tight constraints imposed by such class of microcontrollers. In addition, preliminary results showed that -in the case of very constrained networks- the search procedure can be also executed without the support of GPUs. 

This brief aims to give further insights on the approach to HW-NAS proposed in \cite{garavagno2023hardware}. In particular, the goals are 
\begin{itemize}
\item  to prove empirically that even using a lightweight search procedure the tailored HW-NAS can adapt to the hardware constraints imposed by the target platform; experiments show that the revised procedure can generate custom tiny CNNs that can run in real-time on low-power MCUs. 

\item To show that the search can run on embedded systems; to do this, a constraint on the available memory of the device that executes the NAS procedure is added to the optimization problem.
\end{itemize}

In detail, the generated CNNs achieved state-of-the-art results in the human-recognition tasks on the Visual Wake Word dataset, a standard TinyML benchmark. The HW-NAS procedure ran on a laptop with 16 GB of RAM in less than 4 CPU hours\footnote{code available at https://github.com/AndreaMattiaGaravagno/NanoNAS.}. 

\vspace{-10pt}

%% file: 2.related.tex
\section{Background}\label{sec:background}

\subsection{Related works} \label{subsec:related}
\vspace{-5pt}
In recent years, great efforts have been made to implement CNNs on devices with limited computational capability. Manually designed CNNs for mobile devices \cite{MobileNetV3,Shufflenetv2,EfficientNet} have surged. Then, the focus shifted to approaches that tried to automatize the process \cite{MNASnet,Fbnet}, which gave birth to HW-NAS, i.e., an automated procedure that takes into consideration the available resources of the target hardware, (e.g., the amount of RAM, Flash memory, or FLOPS).

State-of-the-art HW-NAS targeted the high-performance MCUs line of STMicroelectronics: $\mu$NAS \cite{liberis2021munas}, MCUNet \cite{MCUNet} and Micronets \cite{Micronets} . Table \ref{tab:nucleo_boards} summarizes the differences between a high-end MCU and the widespread ultra-low-power line hosted in many sensing nodes. Recently Garavagno et al. \cite{garavagno2024colabnas,garavagno2023hardware,garavagno2024running} addressed the design of a HW-NAS that targets ultra-low-power MCUs. These works showed that the constraints imposed by the ultra-low-power line require an ad-hoc approach that deviates from state-of-the-art HW-NAS. 
\vspace{-5pt}

\subsection{Lightweight HW-NAS: overview} \label{subsec:related}
Three hallmarks characterize a HW-NAS: the search space, the optimization problem, and the search strategy. The HW-NAS presented in \cite{garavagno2023hardware} relies on constrained cell-wise search space, setting the basis for achieving two goals: 1) the ability to generate CNNs that fit hard constraints on the available resources of the target hardware, and 2) the opportunity to properly modulate the computation cost of the search process. The first goal is a prerequisite for a HW-NAS that should deal with ultra-low-power MCUs. The second goal is a prerequisite for a HW-NAS that should run on embedded systems. 

In the adopted search space a candidate solution stems from an established structure which is organized as follows \cite{garavagno2023hardware}:
\begin{enumerate}
\item a pre-processing pipeline performing min-max standardization
and batch normalization on the input data; 
\item a convolutional layer with $k$ kernels; 
\item $c$ cells;
\item a classifier with a global average pooling layer followed by a dropout layer which feeds a final layer having softmax activation and a number of neurons equal to the number of classes.  
\end{enumerate}
Here, the $cell$ is a sequence of three layers: a 2-dimensional max pooling, a batch normalization layer, and a convolutional layer. Every max pooling layer halves the input resolution using a 2x2 receptive field with (2,2) stride. Convolutional layers use (3,3) kernels with (1,1) stride and ReLu activation. The number of kernels $n_c$ used in the $c$th cell depends on $k$ according to the following rule:
\begin{equation} \label{eq:kernels}
\begin{aligned}
n_{c} = \begin{cases}
   \multicolumn{1}{@{}c@{\quad}}{k} & if \quad c = 0\\
   \left \lceil{ (2 - \sum_{i=1} ^{c - 1} 2^{-i}) \cdot n_{c - 1} }\right \rceil & if \quad c \geq 1
\end{cases}
\end{aligned}
\end{equation} 
Thus, in the adopted search space admissible solutions can be described by two parameters: $k$ defines the number of kernels used in the first convolutional layer; $c$ defines the number of cells. 
\vspace{-5pt}

\begin{table}[!t]
    \centering
        \begin{tabular}{l c c c} 
                        \bf{STM32 MCU}     & \bf{RAM [kiB]} & \bf{Flash [kiB]} & \bf{CoreMark}  \tabularnewline \hline
                        \multicolumn{4}{c}{high performance line} \tabularnewline
                        F412ZG        & 256       & 1024        & 608       \tabularnewline 
                        \hline
                        \multicolumn{4}{c}{ultra-low-power line} \tabularnewline 
                        L010RBT6      & 20        & 128         & 75        \tabularnewline 
                        L151UCY6DTR   & 32        & 256         & 93        \tabularnewline 
                        L412KBU3      & 40        & 128         & 273       \tabularnewline
        \end{tabular} 
        \caption{Hardware features of the high-end MCUs targeted by state-of-the-art HW-NAS, versus the MCUs targeted by this work \cite{STM32}.}
        \label{tab:nucleo_boards} 
\end{table}

%% file: 3.proposal.tex
\section{Proposed lightweight HW-NAS}\label{sec:proposal}
The lightweight HW-NAS inherits the overall design from \cite{garavagno2023hardware}. Indeed, enhanced versions of both the optimization problem and the search strategy are adopted. The goal is to combine the capability of generating tiny CNNs that can run in real-time on sensing nodes with the ability of running the NAS itself on an embedded system that may play the role of a central unit in a sensor network.    

\subsection{Optimization Problem}
HW-NAS seeks the best neural architecture for the target hardware in a search space $\mathcal{S}_S$. This translates into a constrained optimization problem where the objective function describes the metric adopted to evaluate candidate solutions (i.e., CNNs) and the constraints set the boundaries of the search. In the proposed problem formulation, a candidate CNN is evaluated by training it and assessing its validation accuracy.  
The constraints for a candidate CNN are the RAM, Flash memory, and multiply and accumulate (MAC) operations available on the target hardware that hosts the inference phase. The latter quantity is used as a valuable proxy for the latency \cite{Micronets}.  In addition, a specific constraint limits the available memory for the training phase on the device hosting the search procedure.

This leads to the problem formulation $P$ 
\begin{equation} \label{eq:problem}
\begin{aligned}
P: \begin{cases}
\max f(x)\\
   \phi_{R}(x) \leq \xi_{R}, \phi_{F}(x) \leq \xi_{F}, \phi_{M}(x) \leq \xi_{M}, \theta(x) \leq \Theta_T   \\
   \xi_{R}, \xi_{F}, \xi_{M}, \Theta_T > 0 
\end{cases}
\end{aligned}
\end{equation}
where $x = (k, c)$. Function $f$ returns the validation accuracy. Parameters $\xi_{R}$, $\xi_{F}$, and $\xi_{M}$ represent, respectively, the upper bounds for RAM usage, Flash usage, and MAC operations on the target hardware; $\Theta_T$ sets the bounds on RAM usage for the search procedure. Thus, function $\phi_{R}$ returns the CNN's RAM occupancy, $\phi_{F}$ returns the CNN's Flash occupancy, and $\phi_{M}$ returns the number of MAC operations required by the CNN, while $\theta$ returns the RAM occupancy on the device hosting the NAS. These quantities depend on the number of kernels $k$ used in the first convolutional layer, the number of cells $c$, the magnitude of the network's input size $vi$, and the adopted platform, in this case TF Lite Micro \cite{david2021tensorflow}. 
The search variable $x = (k, c)$ does not include $vi$, which is a fixed quantity defined by the user. 
\vspace{-5pt}

\subsection{Search Strategy}
The optimization problem $P$ is solved with a derivative-free technique, as $f(x)$ is not differentiable \cite{garavagno2023hardware}. 
This procedure compares candidate networks by assessing their validation score $f(x)$; so, in principle, a full training procedure of each candidate CNN is required. Training, though, is a demanding step; the computation cost of the NAS itself is roughly proportional to the number of training procedures to be completed for implementing the search strategy. Thus, a NAS that should run on resource-constrained devices needs a custom design. 

Algorithm \ref{alg:search} formalizes the proposed search strategy, which relies on three factors: 1) the adopted search space inherently generates tiny CNNs as candidate solutions; 2) early stopping is exploited to limit the computation cost of the training process (Sec. \ref{subsec:early} will give further insight into this aspect); 3) a specific constraint is inserted to take into account the resource availability in the platform that runs the NAS. The search starts with $k=1$ kernels in the first layer. Given $k$, the algorithm increments the number of cells $c$ until the corresponding $CNN(k,c)$ 1) satisfies the constraints imposed by parameters $\xi_{R}$, $\xi_{F}$, and $\xi_{M}$ and 2) scores a validation accuracy $f(k,c)$ larger than that achieved with $CNN(k,c-1)$. Then, the same process is completed with $k+1$. If the best CNN obtained with $k+1$ betters the best CNN obtained with $k$, the search continues by further incrementing $k$. Otherwise, the search stops. 

When the training of a network $CNN(k,c)$ cannot be completed due to limited resources available on the computing system running the NAS, the related accuracy $f(k,c)$ is set to 0. Hence, the search strategy can also run on resource-constrained embedded systems.  

\begin{algorithm} [!t]
\caption{Search Strategy}\label{alg:search}
\begin{algorithmic}
\State $k \gets 1$ \Comment{Minimum number of kernels of the first layer} 
\State $(c,best)$ = $BestCNN(k)$ \Comment{call the procedure}
\Repeat 
\State $max=best$
\State $k \gets k + 1$ \Comment{increase number of kernels}
\State $(c,best)$ = $BestCNN(k)$ \Comment{call the procedure}
\Until {$best>max$} 
\State \textbf{return} $(k,c),max$
\Procedure{BestCNN}{k}
\State $best = 0$, $\Delta = 0$ 
\State $c \gets 0$ \Comment{no cells}
\Repeat 
\State train CNN$(k,c)$ \Comment{train the new candidate}
\State $f(k,c)$ = CNN$(k,c)$ \Comment{validation accuracy} 
\State $\Delta$ = $f(k,c) - best$ \Comment{check if accuracy is increasing}
\State $best$ = $f(k,c)$ 
\State $c \gets c + 1$ \Comment{add a cell}
\Until {CNN$(k,c)$ satisfies (\ref{eq:problem}) AND $\Delta>0$ }
\State \textbf{return} $c,best$
\EndProcedure

\end{algorithmic}
\end{algorithm}

\subsection{Early stop strategy} \label{subsec:early}
CNNs are trained using gradient descent techniques, which require applying the backpropagation algorithm multiple times. This work focuses on tiny networks featuring a small set of parameters. Hence, the representation capability is limited and one cannot exploit standard approaches to speed up the convergence of the training. It is well known, though, that early stopping may play the role of a regularization technique, as, in general, the loss function dramatically decreases in the first few epochs of training. Thus, one can rely on a coarse version of early stopping to find a trade-off between the computational cost of the search strategy and the ability to find an effective setting for the pair $(k,c)$. Such a solution is viable in that the actual goal is not to get a reliable estimation of $f$ itself, but only to adopt a good criterion for deciding if a CNN is more promising than the others. 

Then, considering a full training with $n_{ep}$ epochs, if the estimation is performed after  $n_{ep}'$ epochs the speedup is roughly $n_{ep}/n_{ep}'$. Interestingly, this solution impacts only the time (and energy) without effects on the memory occupation. Obviously, when $n_{ep}>>n_{ep}'$ the speedup is higher but the risk of wrong selections grows coherently. 

Thus, the proposed HW-NAS utilizes early stopping to reduce the computation cost of the search procedure and at the same time to tackle issues such as overfitting and local minima. Sec. \ref{subsec:extreme_early_stopping} will show empirically that one can even stop training after very few epochs without hindering the effectiveness of the NAS. 
\vspace{-5pt}

%% file: 4.exp.tex
\section{Experiments}\label{sec:exp}

The experiments involved three datasets: Visual Wake Words (VWW)\cite{vww}, Cifar-10 (C-10) \cite{krizhevsky2009learning}, and Melanoma Skin Cancer \cite{muhammad_hasnain_javid_2022}.
VWW and C-10 are standard benchmarks for Tiny Visual ML. The target platforms belong to the STM32 ultra-low-power line: L010RBT6 (from now on L0), L151UCY6DTR (L1), and L412KBU3 (L4). The constraints in (\ref{eq:problem}) were set according to Table \ref{tab:nucleo_boards}; the MAC upper bound was obtained by rescaling the CoreMark score of a $10^{4}$ factor. Early stopping ($n_{ep}'=3$) was adopted in the search strategy. The CNN selected by the NAS was then trained for 100 epochs with a batch size of $128$ and a learning rate of $10^{-2}$ using the Adam \cite{Adam} optimizer. The validation set collected $10\%$ of the original training set. Generalization performance was measured on public test sets. 

We measured runtime RAM and Flash using X-CUBE-AI 8.1.0 with Tensorflow Lite Micro. Publicly available models of state-of-the-art networks were downloaded in the TF Lite Micro format. The HW-NAS was executed on a laptop with an 11th Gen Intel\textregistered Core\texttrademark i7-11370H CPU @ 3.30GHz, 16 GB of RAM and 512 GB of SSD. 

\subsection{Evaluation of the hardware-aware feature} \label{sec:hardware_awareness}
The first experiment wanted to prove that the proposed HW-NAS can adapt the generated CNNs to the target hardware, even in the case of ultra-low-power MCUs. Table \ref{tab:hardware_awareness} presents the outcomes of this experiment. The columns of the Table  are divided into four sectors. The first sector refers to the MCUs. The second sector gives the pair $(k,c)$ characterizing the selected CNN and the cost of the NAS search procedure, expressed in time. The third sector shows the features of the selected CNN: RAM occupancy, Flash occupancy, and MAC operations. The last sector gives the performance of the CNN in terms of test accuracy and latency (inference time per image). Latency was assessed using STM32Cube.Ai runtime on L4 in balanced mode.

\begin{table}[!t]    \centering
        \begin{tabular}{l | c c |c c c| c  c} 
                       MCU & \multicolumn{2}{c|}{NAS} & \multicolumn{3}{c|}{Tiny CNN} & \multicolumn{2}{c}{Performance} \tabularnewline

                    \multirow{ 2}{*}{\bf{STM32}} & \bf{Arch.}    & \bf{Cost} & \bf{RAM} & \bf{Flash} & \bf{MAC} & \bf{Acc.} & \bf{Lat.} \tabularnewline 
                         & \bf{(k,c)}    &  \bf{[hh]:[mm]}   & \bf{[kiB]} & \bf{[kiB]} & \bf{[MM]} & \bf{[\%]} & \bf{[mS]}  \tabularnewline \hline

                        \multicolumn{7}{c}{Visual Wake Words ($vi=$ 50x50)} \tabularnewline
                        L0 & (3,3)    & 1:39  & 20     & 10.7    & 0.41  & 71.7 & 56.2 \tabularnewline 
                        L1 & (4,5)    & 2:39  & 24     & 21.1    & 0.66  & 74.1 & 62 \tabularnewline 
                        L4 & (6,4)    & 3:17  & 28.5   & 23.7    & 1.27  & 77 & 87.9      \tabularnewline \hline
                        \multicolumn{7}{c}{Cifar-10 ($vi=$ 32x32)} \tabularnewline 
                        L0 & (6,4)    & 1:07  & 15.5   & 23.9    & 0.54 & 64.8 & 40.1 \tabularnewline 
                        L1 & (8,4)    & 1:15  & 16.5   & 28.8    & 0.7 & 68.5 & 50.3 \tabularnewline 
                        L4 & (14,4)    & 1:46  & 21   & 53.26    & 1.55 & 72.6 & 119.3  \tabularnewline \hline
                        \multicolumn{7}{c}{Melanoma Skin Cancer ($vi=$ 50x50)} \tabularnewline 
                        L0 & (1,3)    & 0:07  & 19.5   & 8.6    & 0.09 & 86.9 & 29.7 \tabularnewline 
                        L1 & (4,4)    & 0:14  & 23   & 16.26    & 0.65 & 88.8 & 61.9 \tabularnewline 
                        L4 & (6,4)    & 0:20  & 28.5 & 23.65    & 1.27 & 91.8 & 87.9   \tabularnewline
        \end{tabular}
            \caption{Performance of the proposed HW-NAS on target MCUs. \vspace{-5pt}}
        \label{tab:hardware_awareness}
\end{table}

The proposed HW-NAS adapted the architecture of the generated CNNs to each platform. The search cost raised as the hardware resources of the target device increased because of the broader search space. The search costs in Table \ref{tab:hardware_awareness} indeed represent a worst case, as they were obtained without involving any GPU. The experiment involving the L4 MCU as a target on the VWW was repeated by executing the NAS on a Raspberry Pi 4 with 4 GB of RAM; the search cost was 34:29 [hh]:[mm]. Such result confirmed that the proposed HW-NAS can also run successfully on resource-constrained devices.  

As expected, the larger CNNs have been generated when the L4 MCU was selected as target. The generalization performance of the selected CNNs again scales with their size. Overall, the proposed method can cope with very hard constraints by balancing the generalization performance with the available hardware resources.

\subsection{State-of-the-art comparison} \label{sec:state_of_the_art_comparison}

Table \ref{tab:sota_comp} presents a comparison on the VWW dataset with three state-of-the-art HW-NAS for MCUs: MCUNet \cite{MCUNet}, Micronets \cite{Micronets}, and ColabNAS \cite{garavagno2024colabnas}. The first two works focus on high-performance MCUs, whereas \cite{garavagno2024colabnas} targets ultra-low-power MCUs. Thus, the comparison involves -for MCUNet and Micronets- the smallest CNN among those presented in the respective works. In the case of MCUNet the target device was a STM32F412 \cite{MCUNet}, while for Micronets and ColabNAS the target was a STM32F446RE \cite{Micronets}. For the proposed HW-NAS the table reports the CNN selected when L4 was the target.     

\begin{table}[t]
    \centering
    \begin{tabular}{cccccc}
       Model  & Acc & RAM & Flash & MACC  & Serch time \\
           & [\%] & [kiB] & [kiB] &[MM] & [hh]:[mm] \\
        \hline

        Proposal & 77 & 28.5 & 23.7 & 1.3 & 3:17 CPU / 1:08 GPU\\ 
        ColabNAS \cite{garavagno2024colabnas} & 77.6  & 31.5 & 20.83 & 2 & 7:09 CPU / 2:28 GPU  \\
        Micronets \cite{Micronets} & 76.8 & 70.5 & 273.8 & 3.3 & n.a GPU\\
        MCUNet  \cite{MCUNet} & 87.4 & 168.5 & 530.5 & 6 & > 300:00 GPU\\
        \hline
    \end{tabular}
    \caption{Comparison with existing HW-NAS on VWW dataset. \vspace{-10pt}}
    \label{tab:sota_comp}
\end{table}

The table shows that the CNN selected by the proposed HW-NAS achieved on the VWW dataset the same accuracy scored by the CNNs selected, respectively, by Micronets \cite{Micronets} and ColabNAS \cite{garavagno2024colabnas}. However, the last two CNNs require more RAM, more Flash memory, and more MAC. In fact, the CNN selected by MCUnet \cite{MCUNet} achieved the highest accuracy. But, again, this CNN cannot be hosted on a ultra-low-power MCU such as the L4. It is worth noting that Lin et al. \cite{MCUNet} showed that a well-known lightweight CNN for image classification tasks, MobileNetV2, requires at least 256 kB of RAM to achieve a satisfactory accuracy on VWW. In fact, such amount of RAM is not available on ultra-low-power MCUs.

The proposed HW-NAS resulted the best in term of search cost. Such outcome confirms the ability to select effective architectures with a  lightweight search strategy. 
\vspace{-5pt}

\subsection{Early stopping: analysis} \label{subsec:extreme_early_stopping}
\begin{figure}[!t]
\centering
\includegraphics[width=3.5in]{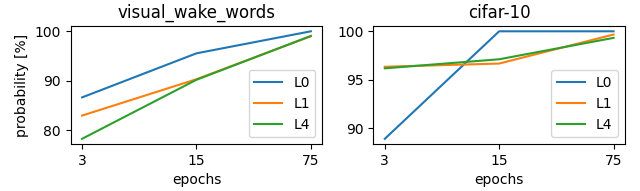}
\caption[caption]{Probability of selecting the best CNN based on the epochs. \vspace{-10pt}}
\label{fig:state_of_the_art_comparison_vww}
\end{figure}

The early stopping criterion is the key factor behind the algorithm's low search cost. Such an approach, though, may affect the ability to find an effective CNN. Figure \ref{fig:state_of_the_art_comparison_vww} shows the probability of selecting the very same CNN singled out using 100 epochs when applying the early stopping criterion with, respectively, 3, 15, and 75 epochs. A training with 100 epochs was used as reference because such setup almost always ensures the convergence to a stable local minima \cite{garavagno2024colabnas}. The plot in Fig. \ref{fig:state_of_the_art_comparison_vww} gives on the x-axis the number of epochs and on the y-axis the probability, which has been assessed empirically. 

The experiment confirms that even using a number of epochs as small as 3, the proposed HW-NAS selects with high probability (> 78\%) the most promising candidate, i.e., the CNN that would have been selected with 100 epochs. This is a major result if one thinks that the search time scales linearly with the number of epochs, providing a theoretically estimated speed-up of 100/3. However, model initialization and the dataset loading procedure, which must be repeated before every training, strongly degrades such speed-up even more in the presence of large datasets, as in this case.

\vspace{-5pt}